\documentclass[runningheads]{llncs}
\usepackage{enumitem,amssymb}
\usepackage{todonotes}
\usepackage{pifont}

\usepackage{url}
\usepackage{color, colortbl}
\usepackage{booktabs}
\usepackage{fixltx2e}
\usepackage{floatrow}  
\usepackage{blindtext}  
\usepackage{array, caption, floatrow, makecell, booktabs}  
\usepackage{multirow}  
\usepackage{rotating}  
\usepackage{color,soul}

\usepackage{fdsymbol}

\usepackage{amsmath}
\usepackage{dsfont}
\usepackage{mathtools}

\usepackage{algorithmic}

\usepackage{amsmath}
\usepackage{xparse}

\ExplSyntaxOn
\NewDocumentCommand{\ucgreek}{m}
 {\str_case:nn { #1 } {
    {A}{\mathrm{A}} {B}{\mathrm{B}} {C}{\Sigma} {D}{\Delta} {E}{\mathrm{E}} 
    {F}{\Phi} {G}{\Gamma} {H}{\mathrm{H}} {I}{\mathrm{I}} {J}{\Theta} {K}{\mathrm{K}} 
    {L}{\Lambda} {M}{\mathrm{M}} {N}{\mathrm{N}} {O}{\mathrm{O}} {P}{\Pi}
    {Q}{\mathrm{X}} {R}{\mathrm{P}} {S}{\Sigma} {T}{\mathrm{T}} {U}{\Upsilon} 
    {W}{\Omega} {X}{\Xi} {Y}{\Psi} {Z}{\mathrm{Z}}
}}
\NewDocumentCommand{\lcgreek}{m}
 {\str_case:nn { #1 }
   {{a}{\alpha} {b}{\beta} {c}{\varsigma} {d}{\delta} 
    {e}{\varepsilon} {f}{\varphi} {g}{\gamma} {h}{\eta} {i}{\iota}
    {j}{\vartheta} {k}{\kappa} {l}{\lambda} {m}{\mu} {n}{\nu} {o}{o}
    {p}{\pi} {q}{\chi} {r}{\rho} {s}{\sigma} {t}{\tau} {u}{\upsilon} 
    {w}{\omega} {x}{\xi} {y}{\psi} {z}{\zeta}
}}
\ExplSyntaxOff

\newcommand{\mathimage}[1]{\mathbf{#1}}

\newcommand{\mathscal}[1]{\lowercase{\textit{#1}}}
\newcommand{\mathtensor}[1]{\mathrm{\uppercase{#1}}}

\newcommand{\mathfunc}[1]{\uppercase{\ucgreek{#1}}} 
\newcommand{\norm}[1]{\left\lVert#1\right\rVert} 

\usepackage{pifont}

\usepackage[normalem]{ulem}

\newcommand{\lightgreentext}[1]{\color[RGB]{44,198,43}{#1}\color{black}}
\newcommand{\lightbluetext}[1]{\color[RGB]{31,120,180}{#1}\color{black}}
\newcommand{\darkredtext}[1]{\color[RGB]{198,44,43}{#1}\color{black}}

\graphicspath{{figures/}}
\newcommand{\projectpage}{\url{https://vios-s.github.io/multiscale-pyag}}

\begin{document}


\title{Self-supervised Multi-scale Consistency\\for Weakly Supervised Segmentation Learning}
\titlerunning{Self-sup. Multi-scale Consistency for Weakly Sup. Segmentation Learning} 
\author{Gabriele Valvano\inst{1, 2} 
        Andrea Leo\inst{1} \and
        Sotirios A. Tsaftaris\inst{2}}
        
\authorrunning{G. Valvano et al.} 
\institute{
    IMT School for Advanced Studies Lucca, Lucca 55100 LU, Italy 
    \and School of Engineering, University of Edinburgh, Edinburgh EH9 3FB, UK
} 



%
\maketitle              

\begin{abstract}
Collecting large-scale medical datasets with fine-grained annotations is time-consuming and requires experts. For this reason, weakly supervised learning aims at optimising machine learning models using weaker forms of annotations, such as scribbles, which are easier and faster to collect. Unfortunately, training with weak labels is challenging and needs regularisation. 
Herein, we introduce a novel self-supervised multi-scale consistency loss, which, coupled with an attention mechanism, encourages the segmentor to learn multi-scale relationships between objects and improves performance. 
We show state-of-the-art performance on several medical and non-medical datasets. 
The code used for the experiments is available at~\projectpage.

\end{abstract}

\section{Introduction}
To lessen the need for large-scale annotated datasets,
researchers have recently explored weaker forms of supervision \cite{kervadec2020bounding,valvano2021learning}, consisting of weak annotations that are easier and faster to collect. 
Unfortunately, weak labels provide lower quality training signals, making it necessary to introduce regularisation to prevent model overfitting. Examples of regularisation are: forcing the model to produce similar predictions for similar inputs \cite{ouali2020overview,valvano2019temporal}, or using prior knowledge about object shape \cite{kervadec2019constrained,zhou2019prior}, intensity \cite{review2016incorporating}, and position \cite{kayhan2020translation}. 

Data-driven shape priors learned by Generative Adversarial Networks (GAN) are popular regularisers \cite{yi2019generative}, exploiting unpaired masks' availability to improve training. Recently, GANs have been used in weakly supervised learning, showing that they can provide training signals for the unlabelled pixels of an image \cite{zhang2020accl}. 
Moreover, multi-scale GANs also provide information on multi-scale relationships among pixels \cite{valvano2021learning}, and can be easily paired with attention mechanisms \cite{valvano2021learning,zhang2019self} to focus on the specific objects and boost performance.
However, GANs can be difficult to optimise, and they require a set of compatible masks for training. Annotated on images from a different data source, these masks must contain annotations for the exact same classes used to train the segmentor. Moreover, the structures to segment must be similar across datasets to limit the risk of covariate shift. For example, there are no guarantees that optimising a multi-scale GAN using masks from a paediatric dataset will not introduce biases in a weakly supervised segmentor meant to segment elderly images.

\begin{figure}[t]
    \centering
    \includegraphics[width=\linewidth]{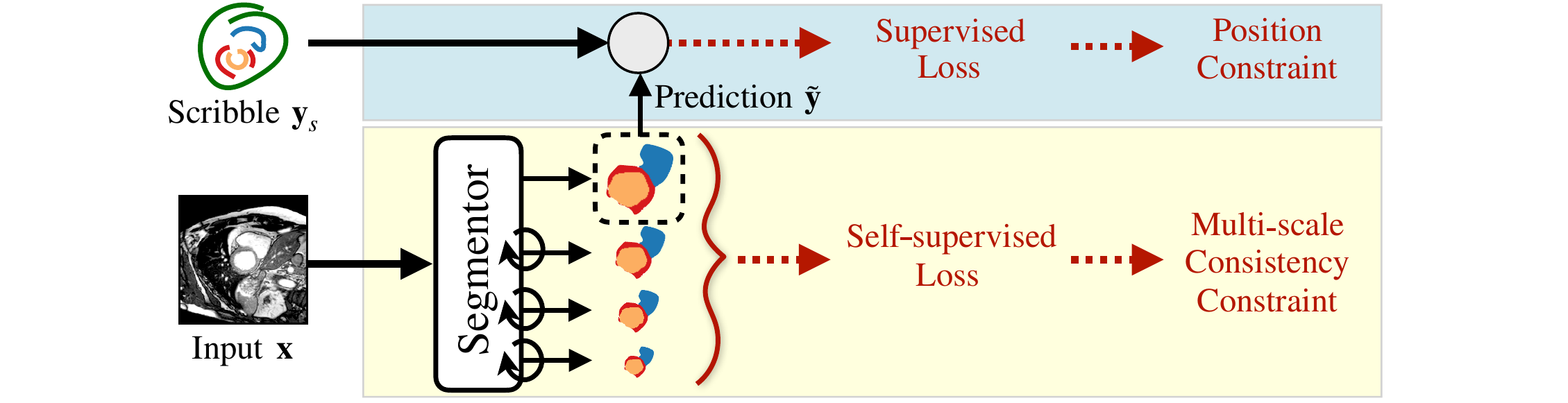}
    \caption{
    We train a segmentor to predict masks overlapping with the available scribble annotations (top of the figure). To encourage the segmentor to learn multi-scale relationships between objects (bottom), we use novel attention mechanism (loopy arrows) that we condition with a self-supervised consistency loss.
    }
    \label{ch7:fig:graphical_abstract}
\end{figure}

Thus, multi-scale GANs are not always a feasible option. In these cases, it would be helpful to introduce multi-scale relationships without relying on unpaired masks. Herein, we show that it is possible to do so without performance loss. Our \textbf{contributions} are: \textbf{i)} we present a novel self-supervised method to introduce multi-scale shape consistency \textit{without} relying on unpaired masks for training; \textbf{ii)} we train a shape-aware segmentor coupling multi-scale predictions and attention mechanisms through a \textit{mask-free} self-supervised objective; and \textbf{iii)} we show comparable performance gains to that of GANs, but without need for unpaired masks. We summarise our idea in Fig.~\ref{ch7:fig:graphical_abstract}.

\section{Related Work}

\noindent\textbf{Weakly-supervised Learning for Image Segmentation.}
Recent research has explored weak annotations to supervise models, including: bounding boxes \cite{kervadec2020bounding}, image-level labels \cite{patel2021weakly}, point clouds \cite{qu2020weakly}, and scribbles \cite{lin2016scribblesup,can2018learning,dorent2020scribble,valvano2021learning}. 
Although it is possible to extend the proposed approach to other types of weak annotations, herein, we focus on scribbles, which have shown to be convenient to collect in medical imaging, especially when annotating nested structures \cite{can2018learning}.

A standard way to improve segmentation with scribbles is to post-process model predictions using Conditional Random Fields (CRFs) \cite{lin2016scribblesup,can2018learning}. 
Recent work avoids the post-processing step and the need of tuning the CRF parameters by including learning constraints during training. For example \cite{belharbi2020deep} uses a max-min uncertainty regulariser to limit the segmentor flexibility, while other approaches regularise training using global statistics, such as the size of the target region \cite{zhou2019prior,kervadec2019constrained,kervadec2020bounding} or topological priors \cite{kervadec2020bounding}. Although they increase model performance, the applicability of these constraints is limited to specific assumptions about the objects and usually requires prior knowledge about the structure to segment. As a result, these methods face difficulty when dealing with pathology or uncommon anatomical variants.
On the contrary, we do not make strong assumptions: we use a general self-supervised regularisation loss, optimising the segmentor to maintain multi-scale structural consistency in the predicted masks.

\noindent\textbf{Multi-scale Consistency and Attention.}
Multi-scale consistency is not new to medical image segmentation. For example, deep supervision uses undersampled ground-truth segmentations to supervise a segmentor at multiple resolution levels \cite{dou20173d}. Unfortunately, differently from these methods, we cannot afford to undersample the available ground-truth annotations because scribbles, which have thin structures, would risk to disappear at lower scales. 

Other methods introduce the shape prior training GAN discriminators with a set of compatible segmentation masks \cite{zhang2020accl,valvano2021learning}. Instead, we remove the need of full masks for training, and we impose multi-scale consistent predictions through an architectural bias localised inside of attention gates within the segmentor.

Attention has been widely adopted in deep learning \cite{Jetley2018} as it suppresses the irrelevant or ambiguous information in the feature maps. 
Recently, attention was also successfully used in image segmentation \cite{oktay2018attention,schlemper2019attention,sinha2020multi}. 
While standard approaches do not explicitly constrain the learned attention maps, Valvano et al. \cite{valvano2021learning} have recently shown that conditioning the attention maps to be semantic increases model performance. 
In particular, they condition the attention maps through an adversarial mask discriminator, which requires a set of unpaired masks to work. 
Herein, we replace the mask discriminator with a more straightforward and general self-supervised consistency objective, obtaining attention maps coherent with the segmentor predictions at multiple scales.

\noindent\textbf{Self-supervised Learning for Medical Image Seg\-men\-ta\-tion.}
Self-su\-per\-vi\-sed learning studies how to create supervisory signals from data using pretext tasks: i.e. easy surrogate objectives aimed at reducing human intervention requirements. 
Several pretext tasks have been proposed in the literature, including image in/out-painting \cite{zhou2019models}, superpixel segmentation \cite{ouyang2020self}, coordinate prediction \cite{bai2019self}, context restoration \cite{chen2019self} and contrastive learning \cite{chaitanya2020contrastive}. 
After a self-supervised training phase, these models need a second-stage fine-tuning on the segmentation task. Unfortunately, choosing a proper pretext task is not trivial, and pre-trained features may not generalise well if unrelated to the final objective \cite{zamir2018taskonomy}. 
Hence, our method is more similar to those using self-supervision to regularise training, such as using transformation consistency \cite{xie2020pgl} and feature prediction \cite{valvano2019temporal}. 

\section{Proposed Approach}\label{ch7:sec:method}

\noindent\textbf{Notation.} 
We use capital Greek letters to denote functions $\mathfunc{f}(\cdot)$, and italic lowercase letters for scalars $\mathscal{s}$. Bold lowercase define two-dimensional images $\mathimage{x} \in \mathds{R}^{h \times w}$, with $h, w \in \mathbb{N}$ natural numbers denoting image height and width. Lastly, we denote tensors $\mathtensor{T} \in \mathds{R}^{n \times m \times o}$ using uppercase letters, with $n, m, o \in \mathbb{N}$. 

\noindent\textbf{Method Overview.} 
We assume to have access to pairs of images $\mathimage{x}$ and their weak annotations $\mathimage{y_{s}}$ (in our case, $\mathimage{y_{s}}$ are scribbles), which we denote with the tuples $(\mathimage{x}, \mathimage{y_{s}})$. 
We present a segmentor incorporating a multi-scale prior learned in a self-supervised manner. We introduce the shape-prior through a specialised attention gate residing at several abstraction levels of the segmentor. These gates produce segmentation masks as an auxiliary task, allowing them to construct semantic attention maps used to suppress background activations in the extracted features.
As our model predicts and refines the segmentation at multiple scales, we refer to these attention modules as Pyramid Attention Gates (PyAG).

\begin{figure}[t]
    \centering
    \includegraphics[width=0.9\linewidth]{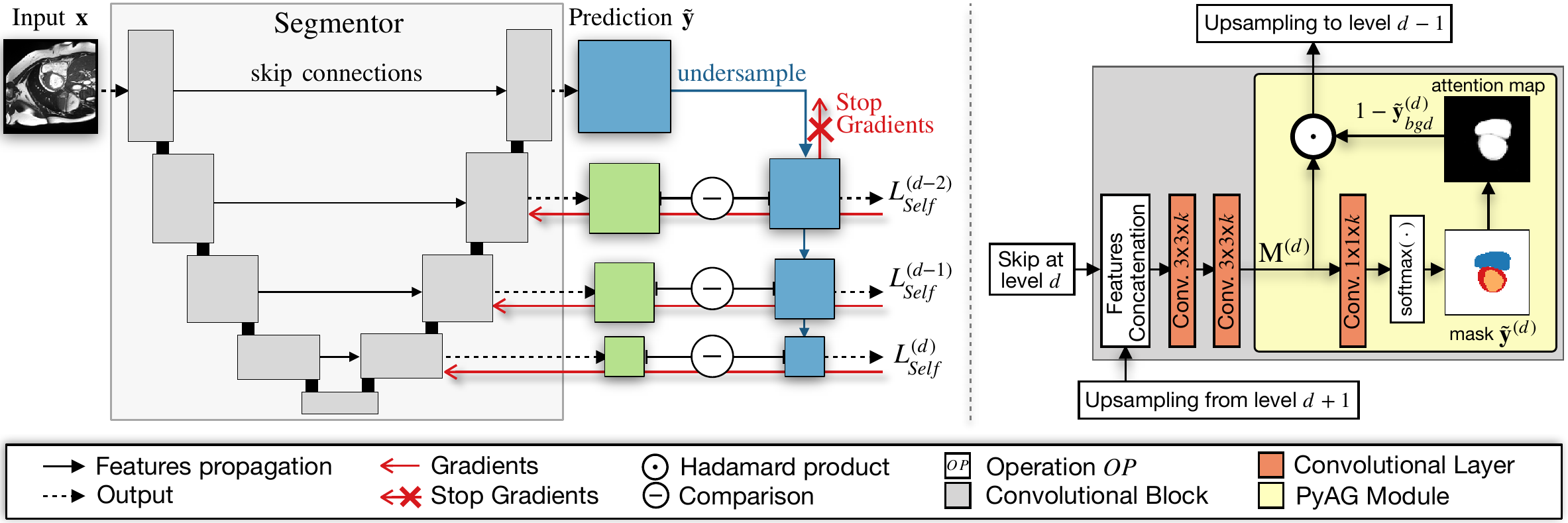}
    \caption{
    \textbf{Left:}
    As a side product of PyAG modules, the segmentor produces segmentation masks at multiple scales. We compare ($\circleddash$ symbol) the lower resolution masks (\lightgreentext{green squares}) to those obtained undersampling the full resolution prediction $\tilde{\mathimage{y}}$ (\lightbluetext{blue squares}), computing a self-supervised loss contribution $\mathcal{L}_{Self}^{(\cdot)}$ at each level. To prevent trivial solutions, we stop (\darkredtext{\textbf{X} } symbol) gradients (\darkredtext{red arrows}) from propagating through the highest resolution stream.
    \textbf{Right:}
    At every depth level $\mathscal{d}$, a convolutional block processes the input features and predicts a low-resolution version of the segmentation mask $\mathimage{y}^{(\mathscal{d})}$ as part of a PyAG attention module (represented in light yellow background). To ensure that the mask $\tilde{\mathimage{y}}^{(\mathscal{d})}$ is consistent with the final prediction $\tilde{\mathimage{y}}$, we use the self-supervised multi-scale loss described in Eq.~\ref{ch7:eq:l_self} and graphically represented on the left panel. Using the predicted mask, we compute the probability of pixels belonging to the background and then suppress their activations in the feature map $\mathtensor{M}^{(\mathscal{d})}$ according to Eq.~\ref{ch7:eq:bkd_masking}.
    }
    \label{ch7:fig:self_sup_and_pyag}
\end{figure}

\noindent\textbf{Model Architecture and Training.}
The segmentor $\Sigma(\cdot)$ is a modified UNet \cite{ronneberger2015u} with batch normalisation \cite{ioffe2015batch}. 
Encoder and decoder of the UNet are interconnected through skip connections, which propagate features across convolutional blocks at multiple depth levels $\mathscal{d}$. 
We leave the encoder as in the original framework while we modify the decoder at each level, as illustrated in Fig.~\ref{ch7:fig:self_sup_and_pyag}. 
In particular, we first process the extracted features with two convolutional layers, as in the standard UNet. Next, we refine them with the introduced PyAG module, represented in a light yellow background on the right side of Fig.~\ref{ch7:fig:self_sup_and_pyag}.  
Each PyAG module consists of: classifier, background extraction, and multiplicative gating operation. 
As a classifier, we use a convolutional layer with $\mathscal{c}$ filters having size $1\times1\times \mathscal{k}$, with $\mathscal{c}$ the number of segmentation classes including the background, and $\mathscal{k}$ the number of input channels. 
Obtained an input feature map $\mathtensor{M}^{(\mathscal{d})}$ at depth $\mathscal{d}$, the classifier predicts a multi-channel score map that we pass through a \emph{softmax}. The resulting tensor assigns a probabilistic value between 0 and 1 to each spatial location. We make this tensor a lower-resolution version of the predicted segmentation mask using the self-supervised consistency constraint:
\begin{equation}\label{ch7:eq:l_self}
    \mathcal{L}_{Self}  
            = - \sum\nolimits_{\mathscal{d}=1}^{\mathscal{n}} 
                 \sum\nolimits_{{\mathscal{i}}=1}^{\mathscal{c}} 
                        \mathimage{\tilde{y}}_{\mathscal{i}}^{(0)} \log(\mathimage{\tilde{y}}_{\mathscal{i}}^{(\mathscal{d})}),
\end{equation}
where $\mathscal{d}$ is the depth level, ${\mathscal{i}}$ is an index denoting the class, $\mathimage{\tilde{y}}^{(\mathscal{d})}$ is the prediction at depth $\mathscal{d}$, and $\mathimage{\tilde{y}}^{(0)} = \mathimage{\tilde{y}}$ is the final prediction of the model.\footnote{Here we assume that the predicted $\mathimage{\tilde{y}}$ is a mask, not a scribble. Intuitively, our hypothesis derives from the observation that unlabelled pixels in the image have an intrinsic uncertainty: thus, the segmentor will look for clues in the image (e.g. anatomical edges and colours) to solve the segmentation task. Since eq. \ref{ch7:eq:l_pce} does not limit model flexibility on the unlabelled pixels, we empirically confirm our hypothesis.} Notice that, different from \cite{valvano2021learning}, we condition $\mathimage{\tilde{y}}^{(\mathscal{d})}$ with $\mathcal{L}_{Self}$ rather than a multi-scale discriminator.

To prevent affecting the final prediction, we propagate the self-supervised training gradients only through the attention gates and the segmentor encoder, as we graphically show in Fig.~\ref{ch7:fig:self_sup_and_pyag}, left.
We further constrain the segmentor to reuse the extracted information by suppressing the activations in the spatial locations of the feature map $\mathtensor{M}^{(\mathscal{d})}$ which can be associated with the background (Fig.~\ref{ch7:fig:self_sup_and_pyag}, right). This multiplicative gating operation can be formally defined as:
\begin{equation}\label{ch7:eq:bkd_masking}
    \mathtensor{M}^{(\mathscal{d})} \leftarrow \mathtensor{M}^{(\mathscal{d})} \cdot 
    \big( 
    1 - \mathimage{\tilde{y}}_{bkd}^{(\mathscal{d})} 
    \big),
\end{equation}
where $\mathimage{\tilde{y}}_{bkd}^{(\mathscal{d})}$ is the background channel of the predicted mask at the depth level $\mathscal{d}$.
The extracted features are finally upsampled to the new resolution level $d-1$ and processed by the next convolutional block. 

To supervise the model with scribbles, we use the Partial Cross-Entropy (PCE) loss \cite{tang2018normalized} on the final prediction $\mathimage{\tilde{y}}$. 
By multiplying the cross-entropy with a labelled pixel identifier $\mathds{1}(\mathimage{y_s})$, the PCE avoids loss contribution on the unlabelled pixels.
The role of the masking function $\mathds{1}(\mathimage{y_s})$ is to return 1 for annotated pixels, 0 otherwise. Mathematically, we formulate the weakly-supervised loss as:
\begin{equation}\label{ch7:eq:l_pce}
     \mathcal{L}_{PCE}  
            = \mathds{1}(\mathimage{y_{s}}) \cdot 
                        \big[ - \sum\nolimits_{\mathscal{i}=1}^{\mathscal{c}} 
                        \mathimage{y_s}_i \log(\mathimage{\tilde{y}}_i) \big],
\end{equation}
with $\mathimage{y_{s}}$ the ground truth scribble annotation. 

Considering both weakly-supervised and self-supervised objectives, the overall cost function becomes: $\mathcal{L} = \mathcal{L}_{PCE} + \mathscal{a} \cdot \mathcal{L}_{Self}$, where $\mathscal{a}$ is a scaling factor that balances training between the two costs. 
Similar to \cite{valvano2021learning}, we find beneficial to use a dynamic value for $\mathscal{a}$, which maintains a fixed ratio between supervised and regularisation cost. 
In particular, we set $\mathscal{a}= \mathscal{a}_0 \cdot \frac{\norm{\mathcal{L}_{Self}}}{\norm{\mathcal{L}_{PCE}}}$, where $\mathscal{a}_0 = 0.1$ is meant to give more importance to the supervised objective. We minimise $\mathcal{L}$ using Adam optimiser \cite{kingma2014adam} with a learning rate of 0.0001, and a batch size of 12.

\section{Experiments}
\subsection{Data}

\noindent \textbf{ACDC}~\cite{bernard2018deep} has cardiac MRIs of 100 patients. There are manual segmentations for right ventricle (RV), left ventricle (LV) and myocardium (MYO) at the end-systolic and diastolic cardiac phases. We resample images to the average resolution of 1.51$mm^2$, and crop/pad them to $224\times224$ pixels. We normalise data by removing the patient-specific median and dividing by its interquartile range. 

\noindent \textbf{CHAOS}~\cite{chaos} contains abdominal images from 20 different patients, with manual segmentation of liver, kidneys, and spleen. We test our method on the available T1 in-phase images. We resample images to 1.89$mm^2$ resolution, normalise them in between -1 and 1, and then crop them to $192\times192$ pixel size. 

\noindent \textbf{LVSC}~\cite{suinesiaputra2014collaborative} has cardiac MRIs of 100 subjects, with manual segmentations of left ventricular myocardium (MYO). We resample images to the average resolution of 1.45$mm^2$ and crop/pad them to $224\times224$ pixels. We normalise data by removing the patient-specific median and dividing by its interquartile range.

\noindent \textbf{PPSS}~\cite{luo2013pedestrian} has (non-medical) RGB images of pedestrians with occlusions. Images were obtained from 171 different surveillance videos and cameras. There are manual segmentations for six pedestrian parts: face, hair, arms, legs, upper clothes, and shoes. We resample all the images to the same spatial resolution of the segmentation masks: $80\times160$; then we normalise images in $[0, 1]$ range. 

\noindent\textbf{Scribbles.} 
The above datasets provide fully-annotated masks. To test the advantages of our approach in weakly-supervised learning, we use the manual scribble annotations provided for ACDC in~\cite{valvano2021learning}. For the remaining datasets, we follow the guidelines provided by Valvano et al. \cite{valvano2021learning} to emulate synthetic scribbles using binary erosion operations or random walks inside the segmentation masks.

\noindent\textbf{Setup.} 
We divide ACDC, LVSC, and CHAOS data into groups of 70\%, 15\% and 15\% of patients for train, validation, and test set, respectively. In PPSS, we follow recommendations in~\cite{luo2013pedestrian}, using images from the first 100 cameras to train (90\%) and validate (10\%) our model, the remaining 71 cameras for testing it.

\subsection{Evaluation Protocol}
We compare segmentation performance of our method, termed \textbf{UNet\textsubscript{PyAG}}, to:
\begin{itemize}
    \item \textbf{UNet}: 
    Trained on scribbles using the $\mathcal{L}_{PCE}$ loss \cite{tang2018normalized}.
    
    \item \textbf{UNet\textsubscript{Comp.}}:
    UNet segmentor whose training is regularised with the Compactness loss proposed by \cite{liu2020shape}, which models a generic shape compactness prior and prevents the appearance of scattered false positives/negatives in the generated masks. 
    The compactness prior is mathematically defined as: $\mathcal{L}_{\text{Comp.}}=\frac{P^{2}}{4 \pi A}$, where $P$ is the perimeter length and $A$ is the area of the generated mask.
    As for our method, we dynamically rescale this regularisation term to be 10 times smaller than the supervised cost (Sec. \ref{ch7:sec:method}).
    
    \item \textbf{UNet\textsubscript{CRF}}:
    Lastly, we consider post-processing the previous UNet predictions through CRF to better capture the object boundaries \cite{chen2017deeplab}.\footnote{CRF models the pairwise potentials between pixels using weighted Gaussians, weighting with values $\omega_1$ and $\omega_2$, and parametrising the distributions with the factors $\sigma_\alpha, \sigma_\beta, \sigma_\gamma$. For ACDC and LVSC, we use the cardiac segmentation parameters in \cite{can2018learning}: $(\sigma_\alpha, \sigma_\beta, \sigma_\gamma, \omega_1, \omega_2) = (2, 0.1, 5, 5, 10)$. For CHAOS, we manually tune $(\omega_1, \omega_2) = (0.1, 0.2)$. Finally, for PPSS, we tuned them to: $(\sigma_\alpha, \sigma_\beta, \sigma_\gamma, \omega_1, \omega_2) = (80, 3, 3, 3, 3)$.}
    
\end{itemize}

While our method does not need a set of unpaired masks for training, we also compare with methods which learn the shape prior from masks:
\begin{itemize}
    \item \textbf{UNet\textsubscript{AAG}} \cite{valvano2021learning}:
    The method upon which we build our model by replacing the multi-scale GAN with self-supervision. The subscript AAG stands for Adversarial Attention Gates, which couple adversarial signals and attention.
    
    \item \textbf{DCGAN}:
    We consider a standard GAN, learning the shape prior from unpaired masks. This model is the same as UNet\textsubscript{AAG}, but without attention gates and multi-scale connections between segmentor and discriminator.
    
    \item \textbf{ACCL} \cite{zhang2020accl}: 
    It trains with scribbles using a PatchGAN discriminator to provide adversarial signals, and with the $\mathcal{L}_{PCE}$ \cite{tang2018normalized} on the annotated pixels.
\end{itemize}

We perform 3-fold cross-validation and measure segmentation quality using Dice and IoU scores, and the Hausdorff Distance.
We use Wilcoxon test ($p<0.01$) to show if improvements w.r.t. the second best model are statistically significant.

\subsection{Results}
\begin{figure}[t]
    \centering
    \includegraphics[width=\linewidth]{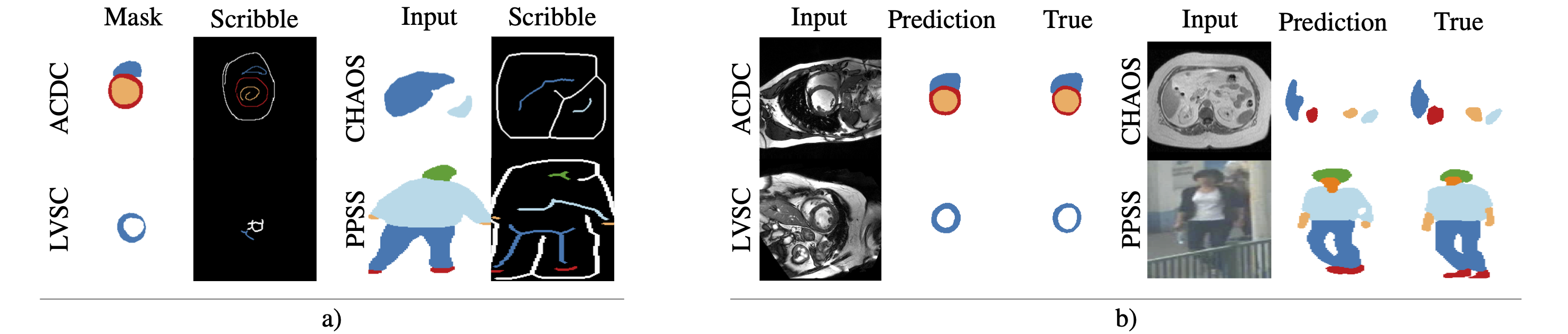}
    \caption{
    For each dataset: a) examples of scribbles b) model predictions.
    }
    \label{ch7:fig:segmentation_panel}
\end{figure}

\begin{figure}[h!]
    \centering
    \includegraphics[width=\linewidth]{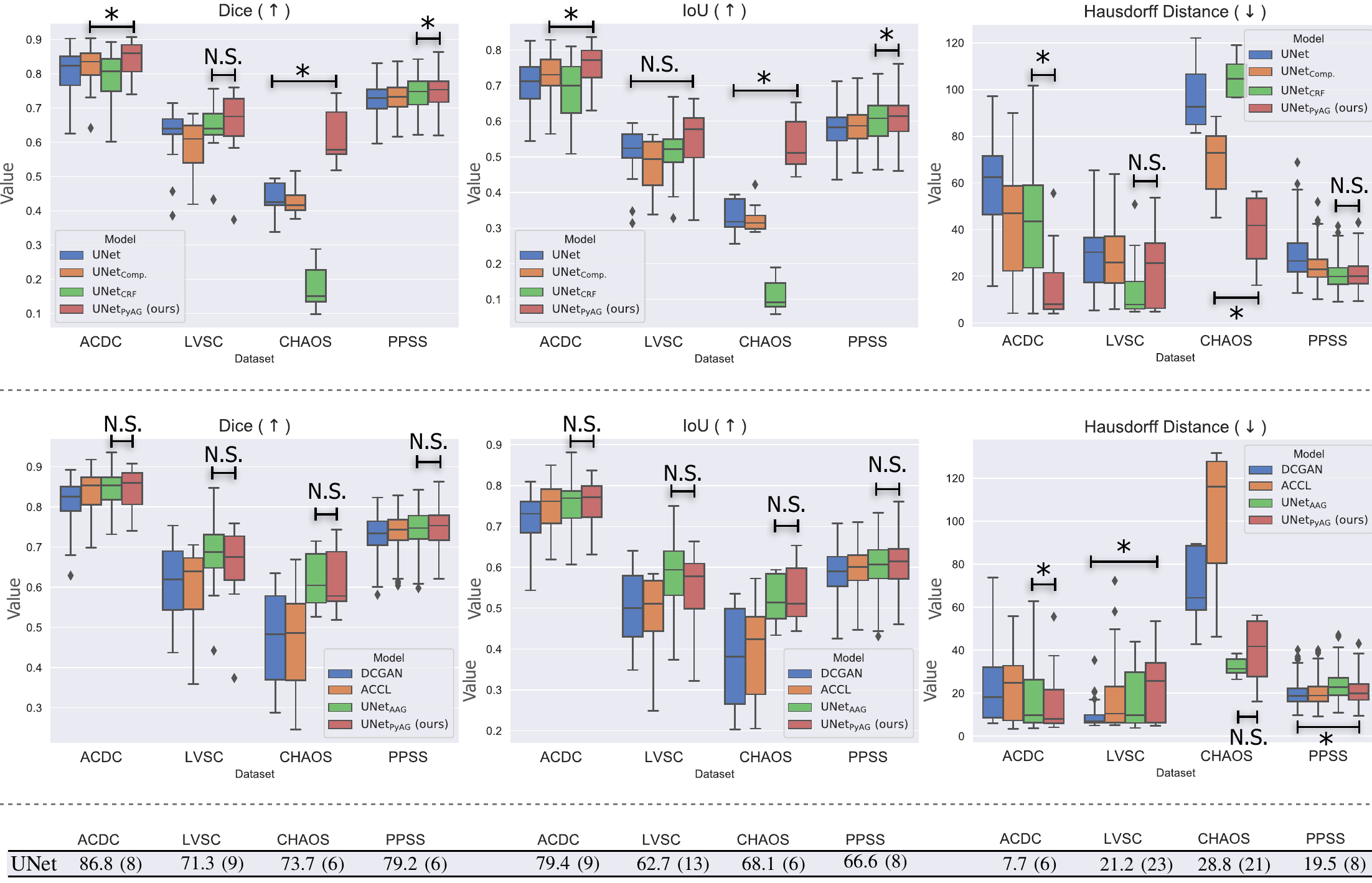}
    \caption{
    Segmentation performance in terms of Dice (\textuparrow), IoU (\textuparrow), Hausdorff distance (\textdownarrow), with arrows showing metric improvement direction. Box plots report median and inter-quartile range (IQR), considering outliers the values outside $2 \times$IQR. 
    For each dataset, we compare the two best performing models (horizontal black lines) and use an asterisk (*) to show if their performance difference is statistically significant (Wilcoxon test, $p<0.01$) or N.S. otherwise.
    \textbf{Top row:} our method vs baseline (UNet) and methods regularising predictions with Compactness loss (UNet\textsubscript{Comp.}) and CRF as post-processing (U\-Net\textsubscript{CRF}). Our method is the best across datasets. In this case, we would like to perform better than the benchmarks ($p<0.01$).
    \textbf{Middle row:} our method vs methods regularising predictions using a shape prior learned from unpaired masks (DCGAN, ACCL, U\-Net\textsubscript{AAG}). In this case, we would like our method to perform at least as well as methods using masks (i.e. we would like the test to be not statistically significant, N.S.). We observe competitive performance with the best benchmark, while we also do not need masks for training.
    \textbf{Bottom row:} we report performance of a UNet trained with fully-annotated masks. These values can be seen as an upper bound when training with scribbles.
    }
    \label{ch7:fig:results_box_plots}
\end{figure}

We show examples of predicted masks in Fig.~\ref{ch7:fig:segmentation_panel} and quantitative results in Fig.~\ref{ch7:fig:results_box_plots}. %

As shown, our method is the best one when we compare it to other approaches that do not require extra masks for training (Fig. \ref{ch7:fig:results_box_plots}, top). In particular, a simple UNet has unsatisfying performance, but regularisation considerably helps. Adding the compactness loss aids more with compact shapes, such as those in ACDC, CHAOS and PPSS, while it can be harmful when dealing with non-compact masks, such as that of the myocardium (doughnut-shape) in LVSC.

Post-processing the segmentor predictions with CRF can lead to performance increase when object boundaries are well defined. On the contrary, we could not make the performance increase on CHAOS data, where using CRF made segmentation worse with all the metrics.

On LVSC, the introduced multi-scale shape consistency prior tends to make the model a bit less conservative on the most apical and basal slices of the cardiac MRI. Unfortunately, whenever there is a predicted mask but the manual segmentation is empty, the Hausdorff distance peaks. 
In fact, by definition, the distance assumes the maximum possible value (i.e. the image dimension) whenever one of the masks is empty, which makes the performance distribution on the test samples broader (see Hausdorff distance box plots for LVSC, Fig. \ref{ch7:fig:results_box_plots}, top).

On CHAOS, Dice and IoU are more skewed for methods not using unpaired masks for training (Fig. \ref{ch7:fig:results_box_plots}, top row). This happens because CHAOS is a small dataset, and optimising models using only scribble supervision is challenging. On the contrary, the extra knowledge of unpaired masks may help (bottom row).

Finally, we compare our method with approaches using unpaired masks for training (Fig. \ref{ch7:fig:results_box_plots}, bottom). We find competitive performance on all datasets. While, in some cases, the UNet\textsubscript{AAG} performs slightly better than UNet\textsubscript{PyAG}, we emphasise that our approach can work also without unpaired masks.

\section{Conclusion}
We introduced a novel self-supervised learning strategy for semantic segmentation. Our approach consists of predicting masks at multiple resolution levels and enforcing multi-scale segmentation consistency. We use these multi-scale predictions as part of attention gating operations, restricting the model to re-use the extracted information on the object shape and position. Our method performs considerably better than other scribble-supervised approaches while having comparable performance to approaches requiring additional unpaired masks to regularise their training. 
Hoping to inspire future research, we release the code used for the experiments at \projectpage.

\subsubsection{Acknowledgments}
This work was partially supported by the Alan Turing Institute (EPSRC grant EP/N510129/1). S.A. Tsaftaris acknowledges the support of Canon Medical and the Royal Academy of Engineering and the Research Chairs and Senior Research Fellowships scheme (grant RCSRF1819\textbackslash8\textbackslash25).

%
\bibliographystyle{splncs04}
\bibliography{references}

\begin{thebibliography}{10}
\providecommand{\url}[1]{\texttt{#1}}
\providecommand{\urlprefix}{URL }
\providecommand{\doi}[1]{https://doi.org/#1}

\bibitem{bai2019self}
Bai, W., Chen, C., Tarroni, G., Duan, J., Guitton, F., Petersen, S.E., Guo, Y.,
  Matthews, P.M., Rueckert, D.: {Self-Supervised Learning For Cardiac MR Image
  Segmentation by Anatomical Position Prediction}. In: MICCAI. Springer (2019)

\bibitem{belharbi2020deep}
Belharbi, S., Rony, J., Dolz, J., Ayed, I.B., McCaffrey, L., Granger, E.: {Deep
  Interpretable Classification and Weakly-Supervised Segmentation of Histology
  Images via Max-Min Uncertainty}. ar\-Xiv preprint arXiv:2011.07221  (2020)

\bibitem{bernard2018deep}
Bernard, O.e.a.: {Deep Learning Techniques for Automatic MRI Cardiac
  Multi-Structures Segmentation and Diagnosis: Is the Problem Solved?} IEEE TMI
   (2018)

\bibitem{can2018learning}
Can, Y.B., Chaitanya, K., Mustafa, B., Koch, L.M., Konukoglu, E., Baumgartner,
  C.F.: {Learning to Segment Medical Images With Scribble-Supervision Alone}.
  In: DLMIA and MLCDS. Springer (2018)

\bibitem{chaitanya2020contrastive}
Chaitanya, K., Erdil, E., Karani, N., Konukoglu, E.: {Contrastive Learning of
  Global and Local Features for Medical Image Segmentation with Limited
  Annotations}. NeurIPS  \textbf{33} (2020)

\bibitem{chen2019self}
Chen, L., Bentley, P., Mori, K., Misawa, K., Fujiwara, M., Rueckert, D.:
  Self-supervised learning for medical image analysis using image context
  restoration. MIA  \textbf{58},  101539 (2019)

\bibitem{chen2017deeplab}
Chen, L.C., Papandreou, G., Kokkinos, I., Murphy, K., Yuille, A.L.: {DeepLab:
  Semantic Image Segmentation With Deep Convolutional Nets, Atrous Convolution,
  and Fully Connected CRFs}. IEEE TPAMI  \textbf{40}(4),  834--848 (2017)

\bibitem{dorent2020scribble}
Dorent, R., Joutard, S., Shapey, J., Bisdas, Sotirios A.nd~Kitchen, N.,
  Bradford, R., Saeed, S., Modat, M., Ourselin, S., Vercauteren, T.:
  {Scribble-based Domain Adaptation via Co-segmentation}. In: MICCAI. pp.
  479--489. Springer (2020)

\bibitem{dou20173d}
Dou, Q., Yu, L., Chen, H., Jin, Y., Yang, X., Qin, J., Heng, P.A.: {3D Deeply
  Supervised Network for Automated Segmentation of Volumetric Medical Images}.
  MIA  \textbf{41},  40--54 (2017)

\bibitem{ioffe2015batch}
Ioffe, S., Szegedy, C.: {Batch Normalization: Accelerating Deep Network
  Training By Reducing Internal Covariate Shift}. In: International Conference
  on Machine Learning (ICML). pp. 448--456. PMLR (2015)

\bibitem{Jetley2018}
Jetley, S., Lord, N.A., Lee, N., Torr, P.H.S.: {Learn To Pay Attention}. ICLR
  (2018)

\bibitem{chaos}
Kavur, A.E., Selver, M.A., Dicle, O., Barış, M., Gezer, N.S.: {CHAOS -
  Combined (CT-MR) Healthy Abdominal Organ Segmentation Challenge Data} (Apr
  2019)

\bibitem{kayhan2020translation}
Kayhan, O.S., Gemert, J.C.v.: {On Translation Invariance in CNNs: Convolutional
  Layers Can Exploit Absolute Spatial Location}. In: CVPR. pp. 14274--14285
  (2020)

\bibitem{kervadec2019constrained}
Kervadec, H., Dolz, J., Tang, M., Granger, E., Boykov, Y., Ayed, I.B.:
  {Constrained-CNN Losses for Weakly Supervised Segmentation}. MIA
  \textbf{54},  88--99 (2019)

\bibitem{kervadec2020bounding}
Kervadec, H., Dolz, J., Wang, S., Granger, E., Ayed, I.B.: {Bounding Boxes for
  Weakly Supervised Segmentation: Global Constraints Get Close to Full
  Supervision}. MIDL  (2020)

\bibitem{kingma2014adam}
Kingma, D.P., Ba, J.: {Adam: A Method for Stochastic Optimization}. ICLR
  (2015)

\bibitem{lin2016scribblesup}
Lin, D., Dai, J., Jia, J., He, K., Sun, J.: {ScribbleSup: Scribble-supervised
  Convolutional Networks for Semantic Segmentation}. In: CVPR. pp. 3159--3167
  (2016)

\bibitem{liu2020shape}
Liu, Q., Dou, Q., Heng, P.A.: {Shape-Aware Meta-Learning For Generalizing
  Prostate MRI Segmentation to Unseen Domains}. In: MICCAI. Springer (2020)

\bibitem{luo2013pedestrian}
Luo, P., Wang, X., Tang, X.: {Pedestrian Parsing Via Deep Decompositional
  Network}. In: ICCV. pp. 2648--2655 (2013)

\bibitem{review2016incorporating}
Nosrati, M.S., Ha\-marneh, G.: {Incorporating Prior Knowledge in Medical Image
  Segmentation: A Survey}. ar\-Xiv pre\-print arXiv:1607.01092  (2016)

\bibitem{oktay2018attention}
Oktay, O., Schlemper, J., Folgoc, L.L., Lee, M., Heinrich, M., et~al.:
  {Attention U-net: Learning Where to Look For the Pancreas}. MIDL  (2018)

\bibitem{ouali2020overview}
Ouali, Y., Hudelot, C., Tami, M.: {An Overview of Deep Semi-Supervised
  Learning}. ar\-Xiv preprint arXiv:2006.052 78  (2020)

\bibitem{ouyang2020self}
Ouyang, C., Biffi, C., Chen, C., Kart, T., Qiu, H., Rueckert, D.:
  {Self-Supervision With Superpixels: Training Few-Shot Medical Image
  Segmentation Without Annotation}. In: ECCV. pp. 762--780. Springer (2020)

\bibitem{patel2021weakly}
Patel, G., Dolz, J.: {Weakly Supervised Segmentation With Cross-Modality
  Equivariant Constraints}. arXiv preprint arXiv: 2104.02488  (2021)

\bibitem{qu2020weakly}
Qu, H., Wu, P., Huang, Q., Yi, J., Yan, Z., Li, K., Riedlinger, G.M., De, S.,
  Zhang, S., Metaxas, D.N.: {Weakly Supervised Deep Nuclei Segmentation Using
  Partial Points Annotation in Histopathology Images}. IEEE TMI
  \textbf{39}(11),  3655--3666 (2020)

\bibitem{ronneberger2015u}
Ronneberger, O., Fischer, P., Brox, T.: {U-net: Convolutional Networks for
  Biomedical Image Segmentation}. In: MICCAI. pp. 234--241. Springer (2015)

\bibitem{schlemper2019attention}
Schlemper, J., Oktay, O., Schaap, M., Heinrich, M., Kainz, B., Glocker, B.,
  Rueckert, D.: {Attention Gated Networks: Learning to Leverage Salient Regions
  in Medical Images}. MIA  \textbf{53},  197--207 (2019)

\bibitem{sinha2020multi}
Sinha, A., Dolz, J.: {Multi-Scale Self-Guided Attention for Medical Image
  Segmentation}. IEEE Journal of Biomedical and Health Informatics  (2020)

\bibitem{suinesiaputra2014collaborative}
Suinesiaputra, A., Cowan, B.R., Al-Agamy, A.O., Elattar, M.A., Ayache, N.,
  Fahmy, A.S., et~al.: {A Collaborative Resource to Build Consensus for
  Automated Left Ventricular Segmentation of Cardiac MR Images}. MIA
  \textbf{18}(1),  50--62 (2014)

\bibitem{tang2018normalized}
Tang, M., Djelouah, A., Perazzi, F., Boykov, Y., Schroers, C.: {Normalized Cut
  Loss for Weakly-Supervised CNN Segmentation}. In: CVPR. pp. 1818--1827 (2018)

\bibitem{valvano2019temporal}
Valvano, G., Chartsias, A., Leo, A., Tsaftaris, S.A.: {Temporal Consistency
  Objectives Regularize the Learning Of Disentangled Representations}. In: DART
  (2019)

\bibitem{valvano2021learning}
Valvano, G., Leo, A., Tsaftaris, S.A.: {Learning to Segment From Scribbles
  Using Multi-Scale Adversarial Attention Gates}. IEEE TMI  (2021)

\bibitem{xie2020pgl}
Xie, Y., Zhang, J., Liao, Z., Xia, Y., Shen, C.: {PGL: Prior-Guided Local
  Self-Supervised Learning for 3D Medical Image Segmentation}. ar\-Xiv preprint
  arXiv: 2011.12640  (2020)

\bibitem{yi2019generative}
Yi, X., Walia, E., Babyn, P.: {Generative Adversarial Network in Medical
  Imaging: A Review}. MIA  \textbf{58},  101552 (2019)

\bibitem{zamir2018taskonomy}
Zamir, A.R., Sax, A., Shen, W., Guibas, L.J., Malik, J., Savarese, S.:
  {Taskonomy: Disentangling Task Transfer Learning}. In: CVPR. pp. 3712--3722
  (2018)

\bibitem{zhang2019self}
Zhang, H., Goodfellow, I., Metaxas, D., Odena, A.: {Self-attention Generative
  Adversarial Networks}. In: ICLR. pp. 7354--7363. PMLR (2019)

\bibitem{zhang2020accl}
Zhang, P., Zhong, Y., Li, X.: {ACCL: Adversarial Constrained-CNN Loss for
  Weakly Supervised Medical Image Segmentation}. ar\-Xiv:\-2005.00328  (2020)

\bibitem{zhou2019prior}
Zhou, Y., Li, Z., Bai, S., Wang, C., Chen, X., Han, M., Fishman, E., Yuille,
  A.L.: {Prior-Aware Neural Network for Partially-Super\-vi\-sed Multi-Organ
  Segmentation}. In: ICCV. pp. 10672--10681 (2019)

\bibitem{zhou2019models}
Zhou, Z., Sodha, V., Siddiquee, M.M.R., Feng, R., Tajbakhsh, N., Gotway, M.B.,
  Liang, J.: {Models Genesis: Generic Autodidactic Models for 3D Medical Image
  Analysis}. In: MICCAI. pp. 384--393. Springer (2019)

\end{thebibliography}
\end{document}